\def\year{2021}\relax
\UseRawInputEncoding
\documentclass[letterpaper]{article} % DO NOT CHANGE THIS
\usepackage{aaai21}  % DO NOT CHANGE THIS
\usepackage{times}  % DO NOT CHANGE THIS
\usepackage{helvet} % DO NOT CHANGE THIS
\usepackage{courier}  % DO NOT CHANGE THIS
\usepackage[hyphens]{url}  % DO NOT CHANGE THIS
\usepackage{graphicx} % DO NOT CHANGE THIS
\usepackage{natbib}  % DO NOT CHANGE THIS AND DO NOT ADD ANY OPTIONS TO IT
\usepackage{caption} % DO NOT CHANGE THIS AND DO NOT ADD ANY OPTIONS TO IT
\usepackage[ruled]{algorithm2e}
\usepackage{amsmath}
\usepackage{amsfonts}
\usepackage{multirow}
\usepackage{multicol}
\usepackage{booktabs}
\usepackage{tabularx}
\usepackage{bigstrut}
\usepackage[switch]{lineno}  %
\nocopyright

\UseRawInputEncoding
\title{Know Deeper: Knowledge-Conversation Cyclic Utilization Mechanism for Open-domain Dialogue Generation}

\author{
    Yajing Sun\textsuperscript{\rm 1,2}, Yue Hu\textsuperscript{\rm 1,2}\footnotemark[1], Luxi Xing\textsuperscript{\rm 1,2}, Jing Yu\textsuperscript{\rm 1,2}, Yuqiang Xie\textsuperscript{\rm 1,2},
    Xiangpeng Wei\textsuperscript{\rm 1,2} \\ 
}
\affiliations{
    \textsuperscript{\rm 1}Institute of Information Engineering, Chinese Academy of Sciences, Beijing, China\\
    \textsuperscript{\rm 2}School of Cyber Security, University of Chinese Academy of Sciences, Beijing, China\\ 
    \{sunyajing, huyue, xingluxi, yujing02, Xieyuqiang, Weixiangpeng\}@iie.ac.cn
}
\begin{document}

\maketitle

\begin{abstract}
    End-to-End intelligent neural dialogue systems suffer from the problems of generating inconsistent and repetitive responses. 
    Existing dialogue models pay attention to unilaterally incorporating personal knowledge into the dialog while ignoring the fact that incorporating the personality-related conversation information into personal knowledge taken as the bilateral information flow boosts the quality of the subsequent conversation.
    Besides, it is indispensable to control personal knowledge utilization over the conversation level. 
    In this paper, we propose a conversation-adaption multi-view persona aware response generation model that aims at enhancing conversation consistency and alleviating the repetition from two folds. 
    First, we consider conversation consistency from multiple views. 
    From the view of the persona profile, we design a novel interaction module that not only iteratively incorporates personalized knowledge into each turn conversation but also captures the personality-related information from conversation to enhance personalized knowledge semantic representation. From the view of speaking style, we introduce the speaking style vector and feed it into the decoder to keep speaking style consistency. To avoid conversation repetition, we devise a coverage mechanism to keep track of the activation of personal knowledge utilization. Experiments on both automatic and human evaluation verify the superiority of our model over previous models.
\end{abstract}
\begin{figure*}[htb]
	\centering
	\includegraphics[width=.80\textwidth]{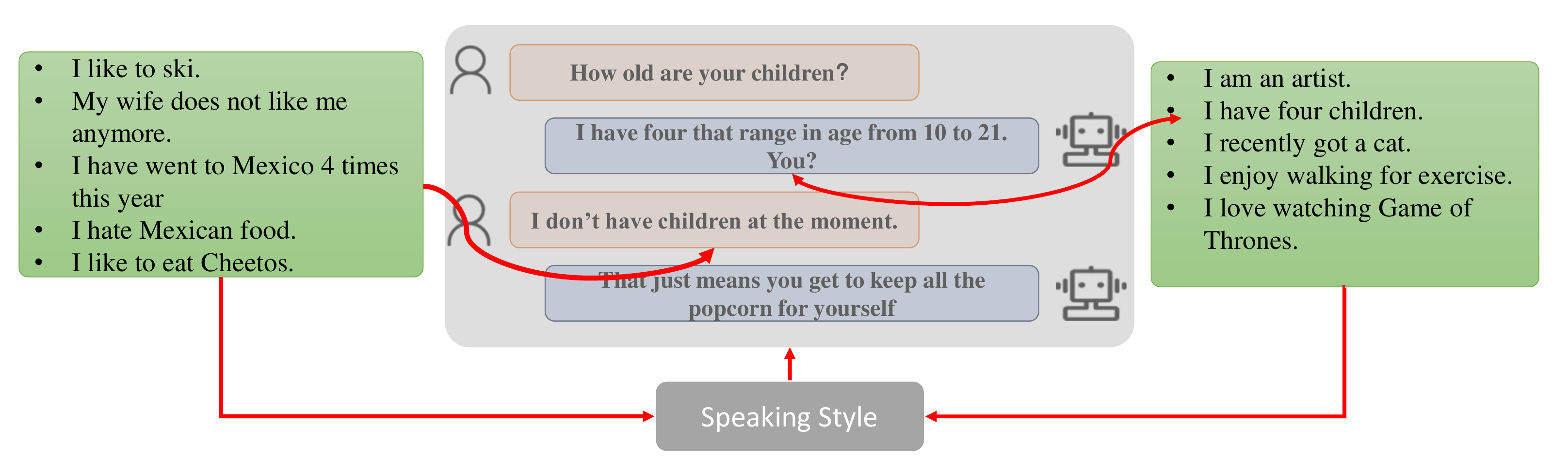}
	\caption{Conversation example with both persona profiles. The green boxes represent the personalized information in the form of text corresponding to humans and machines, respectively. The gray box indicates the content of the conversation. Two-way arrows indicate that conversations and personalized information are mutually reinforcing. Yellow arrows indicate different speaking style between people with different personalities, which also affects the content of the dialogue.}
	\label{introduction}
\end{figure*}
\section{Introduction}
\label{intro}

Generating human-level conversation by machine has been a long-standing goal of artificial intelligence. Because of a large amount of human conversational data available, sequence-to-sequence models and their extensions \cite{sutskever2014sequence,bahdanau2014neural,DBLP:journals/corr/VinyalsL15,shang2015neural} have been widely adopted to learn a generative conversational model. These models incorporate rich knowledge information into the dialogue context and generate diversity responses. Despite great progress has been made, common issues still exist in dialogue systems. First, the dialogue system is incapable of adapting to users with different personalities, which leads to dialogue inconsistent. Second, the module tends to generate repetitive but meaningless content \cite{DBLP:journals/corr/SerbanLCP16} because the same external knowledge is incorporated and utilized multiple times \cite{bao2019know}.

%the repetitiveness of response, which incorporates and utilizes the same external knowledge multiple times to generate the same but meaningless content, remains without a de facto solution.

Presenting a consistent persona is beneficial to gain trust from the users and make users engaged for dialogue system. In recent years, several approaches have been developed to generate consistent responses. There are two methods to construct personalized neural conversation models. \citet{li2016deep} represents user persona knowledge as a vector to capture the speaking style of the speaker implicitly and feed it into the decoder. It's expensive to train these models because they need large quantities of conversational data labeled by user identifiers. Thus, some personal models generate personality-coherent responses using explicit personal profiles through either structural \cite{DBLP:journals/corr/abs-1902-04911} or textual \cite{DBLP:journals/corr/abs-1801-07243}. It’s a matter of fact that persona information, which presents two aspects: persona-profile consistency and speaking style consistency, can greatly improve the consistency and interactivity of the dialogue. 

However most of the existing methods tend to pay attention to utilizing unilateral information that the personal knowledge flows to the conversation to select and incorporate personal knowledge into the dialogue. In fact, leveraging the bilateral information flow of conversation and personal knowledge is of great importance to promote conversation quality. Figure \ref{introduction} illustrates the reasons with an example. First, at the second turn, since the model incorporates personal knowledge \emph{I have four children} into the dialogue, the response is personality-consistent. Second, the personality-related information \emph{from 10 to 21} in the conversation is of great importance for broadening personal knowledge, which is beneficial to subsequent conversations. Moreover, we can see that for the $4$-th turn dialogue, the conversation history is greatly related to the topic \emph{children}. If model is lacking of the effective control over the proper knowledge utilization on the whole conversation level, it's easy to select \emph{I have four children} again, which leads to repetition and couldn't widen more abundant topics to make the conversation more interactive. Intuitively, we can conclude that: (1) The quality of dialogue systems can be improved if the model considers the information interaction between personal knowledge and conversation. (2) It's also indispensable to keep track of the activation of each personal information and coordinate the balance of semantic relevance and repetition. (3) The partner's persona profiles also reflect the model's response generation. 

As a result, we propose the serialized persona aware response generation model to address personality inconsistency and repetition problems. We address conversation consistency from two views. On the one hand, We design a novel interaction module that not only considers to fuse personalized knowledge into the conversation but also captures personality-related information from the conversation to enhance personalized knowledge semantic representation, which results in the success of the subsequent conversation. Moreover, we also keep track of the activation of personal information to avoid repetition. On the other hand, we consider the speaking style based on both speakers' persona profiles and fed it into the decoder to generate a response with profile and speaking style consistent.

The contribution of this work are summarized as follows:
\begin{itemize}
    \item With the object of keeping dialogue consistent, we divide conversation consistency into the profile consistency and speaking style consistency. We model the bilateral semantic information flow between personal knowledge and conversation to keep profile consistency. And we devise the speaking style vector and incorporate it into the decoder to maintain speaking style consistency.
    \item To avoid conversation repetition, we introduce a coverage mechanism to keep track of the activation of knowledge utilization to balance semantic relevance and repetition between conversation and knowledge when we incorporate personal knowledge into the every turn conversation. 
    \item Intensive and extensive experiments have been carried out on ConvAI2 and CMUDoG datasets. The comprehensive experiments demonstrate that our model significantly outperforms the existing methods in keeping dialogue consistency and alleviating repetition.  
\end{itemize}
\begin{figure*}[htb]
	\centering
	\includegraphics[width=0.95\textwidth]{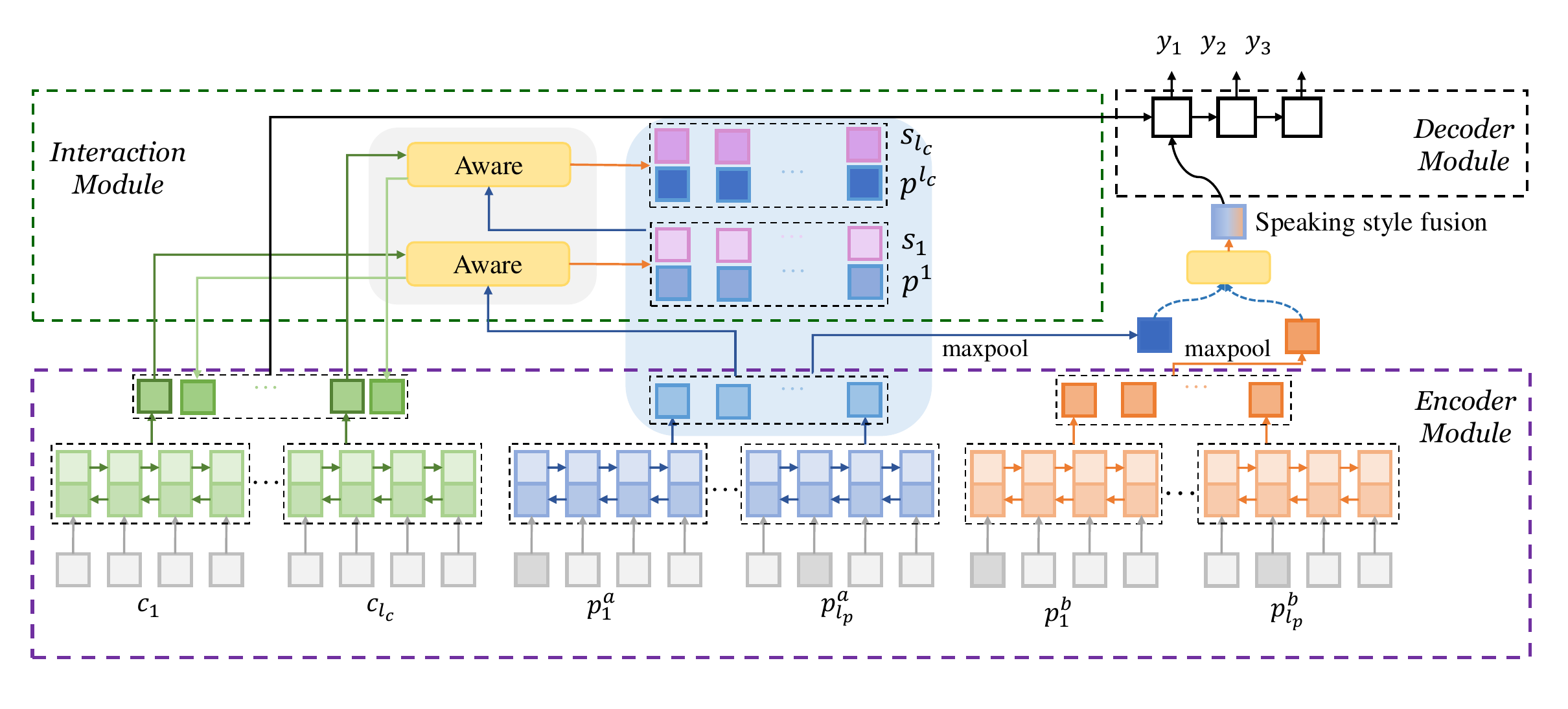}
	\caption{Model Overview. There are three parts including the encoder module, the interaction module, and the decoder module. The network takes context C, two speakers' knowledge sentences $P^A$ and $P^B$ as inputs and generate appropriate responses. In the each step of interaction module, the module will update persona A information $p_1$ and generate coverage vector $s$, which is used to avoid repetition. 
}
	\label{model}
\end{figure*}

\section{Related Work}
Sequence-to-sequence models and their extensions \cite{sutskever2014sequence,bahdanau2014neural} have been widely adopted to learn a generative conversational model from large-scale social conversation data. In recent years, modeling personality consistent dialogue system is drawing increasing attention. The first attempt to model persona is \citeauthor{li2016deep}\shortcite{li2016deep}, which uses the learned persona embedding to capture the users’ background information and speaking style into the model to keep consistency. \citeauthor{DBLP:journals/corr/abs-1710-07388} \shortcite{DBLP:journals/corr/abs-1710-07388} integrates participant role and context information into LSTM. These work crucially depends on the availability of large amounts of speaker-specific conversational data, which are expensive and can't be obtained in many domains. Besides, there are some interesting researchers attempting to use multi-task or transfer learning to model personalized dialogue system. \citeauthor{DBLP:journals/corr/abs-1710-07388} \shortcite{DBLP:journals/corr/abs-1710-07388} proposed a multi-task learning approach. The model utilizes both conversation data across speakers and other types of data pertaining to the speaker and speaker roles to be modeled. Moreover, another methods, namely explicit personalization approaches, attempt to endow dialogue models with persona which is described by natural language sentences or triples. \citeauthor{DBLP:journals/corr/QianHZXZ17} \shortcite{DBLP:journals/corr/QianHZXZ17} constructed the structural personality knowledge and assigned a desired identity to a chat-bot. \citeauthor{DBLP:journals/corr/abs-1801-07243} \shortcite{DBLP:journals/corr/abs-1801-07243} contributed the Persona-Chat dataset which gave a text-described persona, and they further proposed both ranking and generative models. \citeauthor{DBLP:journals/corr/abs-1902-04911} \shortcite{DBLP:journals/corr/abs-1902-04911} pay attention to select appropriate profile knowledge.
%, which use a multiplicative matrix to model different word distribution for different roles. 
 %\citeauthor{zhang-etal-2013-transfer} \shortcite{zhang-etal-2013-transfer} applied domain adaption to the dialogue model and propose a two-phase approach, namely initialization then adaptation and generate personalized responses. 
\citeauthor{DBLP:journals/corr/SerbanLCP16}\shortcite{DBLP:journals/corr/SerbanLCP16} presented the repetitive problems in dialogue systems. \citeauthor{li2016deep}\shortcite{li2016deep} used reinforcement learning methods to model future reward that display the conversational property of non-repetitive turns. \citeauthor{see2019makes}\shortcite{see2019makes} proposed that a good conversation need to avoid repetition, make sense, keeping fluent and coordinate them well. And they define five n-gram based decoding featues to identify repeating bigram features and content words. \citeauthor{bao2019know} \shortcite{bao2019know} proposed Generation-Evaluation framework to control knowledge selection via reinforcement learning. 

But these methods just attempt to incorporate personality information to keep consistency. None of existing models pay attention to incorporating personality-related information of the conversation into the personal knowledge, which benefits the success of the subsequent conversation. So our work focuses on building a multi-turn dialogue system by modeling the bilateral information flow of conversation and personal knowledge to keep conversation consistency.

\section{Background}
Our sentence encoder module is based on the BiGRU with attention mechanism. Specifically, given a sentence $x=\{x_1,x_2,\ldots,x_k\}$ where $x_t \in \mathbb{R}^{d}$ is embedding of $t$-th word in sentence $x$, we firstly to run Bi-GRU \cite{cho-etal-2014-learning} to capture the contextual information. Mathematically, given the word embedding $x_t$ at time step $t$, previous forward hidden vector $\overrightarrow{h_{t-1}}$ and last hidden vector $\overleftarrow{h_{t-1}}$, GRU recurrently computes $h_t$ as follows: 
\begin{equation}
    h_{t}=[\overrightarrow{h_{t}},\overleftarrow{h_{t}}]=[GRU(x_t,\overrightarrow{h_{t-1}});GRU(x_t,\overleftarrow{h_{t-1}})]
\end{equation}
Then we apply attention mechanism to calculate sentence-level contextual representation, which focus more on important semantic information in the sentence-level. Practically, for sentence $h=\{h_1,h_2,\ldots,h_k\}$,
\begin{equation}
    e_{j}=v^T tanh(W_h h_{j})
\end{equation}
\begin{equation}
   h^{\prime}=\sum_{j=1}^{k}\frac{\exp e_{j}}{\sum_{n=1}^{k}\exp e_{n}}h_{j}
\end{equation}
, were $h^{\prime} \in \mathbb{R}^{d}$. We denote the whole encoder module as $f_{\textbf{enc}}(\cdot)$.

\section{Model}
Before presenting the model, we first provide the problem formulation. Suppose that we have a dataset $\mathcal{D}=\{(P^A,P^B,C,Y)_i\}^{N}_{i=1}$. Let $P^A=\{p^{a}_{1},p^{a}_{2},\ldots,p^{a}_{l_p}\}$, where $p^{a}_{i}=\{p^{a}_{i,1},p^{a}_{i,2},\ldots,p^{a}_{i,k}\}$ represents the $i$-th piece personal knowledge of speaker $A$. $l_p$ is the number of knowledge, $k$ is the length of a sentence. Similarly, let $P^B=\{p^{b}_{1},p^{b}_{2},\ldots,p^{b}_{l_p}\}$ represent personal knowledge of speaker $B$. And we denote $C=\{c_1,c_2,\ldots,c_{l_c}\}$ as conversation context with utterances $c_i=\{c_{i,1},c_{i,2},\ldots,c_{i,k}\}$. The current question is located in the last turn. $l_c$ is the turn of context. Our goal is to generate consistent and diversity answer $Y=\{y_1,y_2,\ldots,y_k\}$.

\subsection{Model Overview}
The framework is illustrated in Figure \ref{model} and it consists of three components summarized as follows,
\paragraph{Encoder module}encodes the conversation context and personal knowledge into the semantic representation respectively, which aims to capture the important contextual information of them. And the module also calculate speaking style vector, which will be fed into the decoder.
\paragraph{Interaction module} is responsible for incorporating personal knowledge into the conversation context, updating the knowledge state and enhancing semantic representation of knowledge. The module defines coverage vector to keep track of the activation of personal knowledge utilization, which endows the model the capability of avoiding repetition. And the persona-aware history representation guarantees model consistency and updated personal representation extracts personality-related semantic information from conversation, resulting in successful subsequent conversation.
\paragraph{Decoder module} generates consistent and diversity responses based on the persona-aware context vector and the speaking style user vector.
\subsection{Encode Module}
Understanding the context and scene of the conversation is crucial for a dialogue system. The encoder module is responsible for extracting important contextual information from the sentences to enhance dialogue understanding. It consists of two parts: the first is to encode the context into a dense semantic representation using sentence encoder and the second is to calculate speaking style vector based on the personal knowledge of both parties since speaking style consistency is also important.

The sentence encoder consists of a Bi-GRU component and a self-attention component. Specifically, we encode personal knowledge $P^A$, $P^B$ and conversational context $C$ using $f_{\textbf{enc}}(\cdot)$. Mathematically,
\begin{equation}
    h^{a}_i=f_{\textbf{enc}}(p^{a}_i) 
\end{equation}
\begin{equation}
    h^{b}_i=f_{\textbf{enc}}(p^{b}_i) 
\end{equation}
\begin{equation}
    h^c_i=f_{\textbf{enc}}(c_i) 
\end{equation}
And $h^a, h^b, h^c\in\mathbb{R}^{l_p \times d}$.
Considering the importance of speaking style, we implement the speaking style encoder to calculate a user style vector based on both speakers' personal knowledge. Specifically,
\begin{equation}
    h^{a^{\prime}}= maxpooling(h^a)
\end{equation}
\begin{equation}
    h^{b^{\prime}}= maxpooling(h^b)
\end{equation}
\begin{equation}
    h_{sty}= h^{a^{\prime}} W_{sty} h^{b^{\prime}}
\end{equation}
The speaking style vector is then linearly incorporated into the decoder at each step.
\subsection {Interaction Module}
It's indispensable to leverage the bilateral information flow of conversation and personal knowledge. Incorporating personal knowledge into the dialogue promotes the dialogue consistency, and keeping track of the state of knowledge using on the conversational level also can alleviate repetition. Moreover, it's of great importance to fuse the persona-related conversation information into the persona semantic representation, resulting in diversity and consistent subsequent conversation.

Based on this, we design a serializing persona-conversation interaction module which recurrently updates the personal knowledge from utterance level and progressively incorporate it into the history step-by-step. For sake of semantic relevance, we first use attention mechanism to calculate persona-aware history representation and history-aware persona representation at the turn level based on the history and personal knowledge. To avoid repetitive utilization for personal knowledge, we proposed a coverage vector to keep track of the utilization of persona knowledge and design gate mechanism to get persona-aware history at the conversational level, which is sensitive to the knowledge utilization state. Then we fuse the two different granularity history representation to balance the semantic relevance and repetition. Finally we consider the sequential information between different history turns, and use hierarchical recurrent mechanism to calculate the final history representation. 

The module concentrates on the interaction between the personal knowledge and conversation in every turn. Personal knowledge semantic representation $h^{p(t)}$ and state representation $s_t$ is recurrently updated based on the history information $h^c_t$, we define the dynamic update path as follows, 
\begin{equation} 
    \{h^{p(t)},h^c_t,s_t\}\to \{h^{p(t+1)},h^{c^{\prime}}_t,s_{t+1}\}
\end{equation}
We represent the personal knowledge from two aspects: the semantic representation and the utilization state representation. We define the initial knowledge semantic and state representation are $h^{p(1)} \in \mathbb{R}^{l_p \times d}$ and $s_1 \in \mathbb{R}^{l_p}$ respectively. $s_1$ is a zero vector which means knowledge is not used in the beginning and $h^{p(1)}=h^a$ 

Firstly, since dialogue understanding is closely related to the personal knowledge, we get persona-aware history for sake of semantic relevance as follows,
\begin{equation}
    e^{sem}_{t,j} = \textbf{S}(v^{T} tanh(W_b h^{p(t)} + V_b h^{c}_t))
\end{equation}
\begin{equation}
    h^c_{sem(t)} = \sum_{i=1}^{l_p} e^{sem}_{t,i} h^{p(t)}_{i})
\end{equation}

In terms of the repetition, distinct from the semantic relevant persona-aware history representation $h^c_{sem(t)}$, we combine the coverage vector $s_t$. Because $s_t$ record the history of knowledge utilization, it will discourage the attention which has been heavily attended in the past conversation while implicitly push the attention to the less attended personal knowledge.
Mathematically,
\begin{equation}
\label{repetition}
    e^{rep}_{t,j} = \textbf{S}(v^{T} tanh(W_a h^{p(t)} + V_a h^{c}_t + U_a s_t))
\end{equation}
\begin{equation}
    h^c_{rep(t)} = \sum_{i=1}^{l_p} e^{rep}_{t,i} h^{p(t)}_{i})
\end{equation}

Since $s_t$ consider the knowledge utilization in the past conversation, $h^c_{rep(t)}$ can be viewed as the conversation-level persona-aware history semantic representation, while $h^c_{sem(t)}$ means the turn-level persona-aware history representation.

Finally, we take consistency and repetition of the dialogue into consideration and fuse the two granularity representation $h^c_{sem(t)}$ and $h^c_{rep(t)}$. 
\begin{equation}
    h^{c^{\prime}}_t = \sigma(W_{sem} h^c_{sem(t)}  +W_{rep} h^c_{rep(t)})
\end{equation}
At the every turn, we iteratively update coverage vector through accumulate the attention weights generated by Equation (\ref{repetition}), which is straightforward but effective. Formally,
\begin{equation}
    s_{t+1} = s_t + e^{rep}_{t,j}
\end{equation}
Then we incorporate persona-related history information into the personal information to enhance personal information semantic representation.
\begin{equation}
    h^{p(t)^{\prime}} = \sigma\lbrace W_p \lbrack(h^c_t \oplus h^{p(t)});(h^c_t \odot h^{p(t)}) \rbrack + b \rbrace
\end{equation}
\begin{equation}
    h^{p(t+1)} = h^{p(t)} + h^{p(t)^{\prime}}
\end{equation}

Similar to \citeauthor{DBLP:conf/aaai/XingWWHZ18}\shortcite{DBLP:conf/aaai/XingWWHZ18}, we adopt hierarchical recurrent network to capture sequential contextual information from conversational level. Specifically, the history representation $H=\{h_1,h_2,\ldots,h_{l_c}\}$, where $h_i=\{h^c;h^{c^{\prime}}\}$ is fed to a GRU with attention mechanism to pick up important information from the history into a vector. we represent the final history representation as $O$.

\begin{table*}[htb]
\centering
\resizebox{0.85\textwidth}{!}
{\begin{tabular}{|l|l|c|c|c|}
\hline
\multicolumn{1}{|l|}{\multirow{2}{*}{\textbf{Dataset}}} & \multicolumn{1}{l|}{\multirow{2}{*}{\textbf{Model}}} & \multicolumn{3}{c|}{\textbf{Automatic Evaluation}}  \\ \cline{3-5} 
\multicolumn{1}{|c|}{}                         & \multicolumn{1}{c|}{}                       & \textbf{BLEU-1/2}       & \textbf{Distinct-1/2}   & \textbf{Knowledge R/P/F1} \\ \hline
\multirow{3}{*}{\textbf{ConvAI2(original)}}         & Seq2Seq                                     & 0.3844/0.3046  & 0.0052/0.0191  & 0.007/0.0401/0.0115 \\ \cline{2-5} 
                                               & KG-Net                                      &\textbf{0.4264/0.3342}  & 0.0055/0.0241  & \textbf{0.0103/0.0544/0.0167} \\ \cline{2-5} 
                                               & \textbf{Our model}                                   & 0.4118/0.3268  & \textbf{0.0165/0.0708}  & 0.0066/0.0349/0.0107  \\ \hline
\multirow{3}{*}{\textbf{ConvAI2(revised)}}          & Seq2Seq                                     & 0.3572/0.2828  & 0.0051/0.0173  & 0.0025/0.0139/0.0041   \\ \cline{2-5} 
                                               & KG-Net                                      & \textbf{0.4534/0.3569}  & 0.0041/0.0177  & 0.0035/0.0176/0.0057  \\ \cline{2-5} 
                                               & \textbf{Our model}                                   & 0.4163/0.3300  & \textbf{0.0081/0.0320}  & \textbf{0.0041/0.0196/0.0066}   \\ \hline
\multirow{3}{*}{\textbf{CMUDOG}}                      & Seq2Seq                                     & 0.0.2275/0.1822  & 0.0066/0.0193  & 0.0006/0.0165/0.0010   \\ \cline{2-5} 
                                               & KG-Net                                      & \textbf{0.2632/0.2102}  & 0.0145/0.0462  & 0.0020/0.0467/0.0038  \\ \cline{2-5} 
                                               & \textbf{Our model}                                   & 0.2322/0.1884  & \textbf{0.0465/0.2112}  & \textbf{0.0067/0.1155/0.0122}   \\ \hline
%\multirow{3}{*}{CMU\_DoG}                       & Seq2Seq                                     & 0.2421/0.1949  & 0.0118/0.0327  &-/-/   \\ \cline{2-5} 
%                                               & KG-Net                                      & \textbf{0.2764/0.2214}  & 0.0260/0.0810  & 0.0023/0.0456/0.0043 % \\ \cline{2-5} 
%                                               & Our model                                   & 0.2633/0.2125  & \textbf{0.0829/0.3311}  & \textbf{0.0077/0.1311/0.014}   \\  \hline
\end{tabular}
}
\caption{Experimental results of automatic metrics with the different models on the persona-chat data. There are two different settings for ConvAI2 data: conditioned on the speakers given original persona or revised persona that does not have word overlap.}
\label{automated evaluation}
\end{table*}
%\subsection{Loss Function}
\subsection{Decoder Module}
The decoder generates response based on the persona-aware history $O$ and the speaking style user representation $h_{sty}$. we adopt hierarchical gated fusion Unit (HGFU) \cite{DBLP:journals/corr/abs-1902-04911} decoder to incorporate $h_{sty}$ into the response generation. It's consists of three components. the standard GRU calculate the hidden state for the last generated $y_{t-1}$, persona-style GRU encodes the hidden representation for $h_{sty}$ and fusion unit design a gate mechanism to fuse them and produce the hidden state of the decoder at time $t$. Specifically,
\begin{equation}
    s^y_t=GRU(y_{t-1},s_{t-1})
\end{equation}
\begin{equation}
 s^p_t=GRU(h_{sty},s_{t-1})
\end{equation}
\begin{equation}
   s_t=r \odot s^y_t + (1-r) \odot s^p_t
\end{equation}
where $r=\sigma(V^T[\tanh{W_ys^y_t};\tanh{W_ps^p_t}])$.

Then, we generate the next work $y_t$ according to the hidden state $s_t$ as follows:
\begin{equation}
   p(y_t|\{y_1,\ldots,y_t-1\}) = g(y_{t-1},s_t) 
\end{equation}
where $g$ is a nonlinear function that outputs the probability of $y_t$.

Finally, we use the objective of NLL loss is to quantify the difference between the true response and the response generated by model. It minimizes Negative Log-Likelihood (NLL):
\begin{equation}
    \mathcal{L}(\theta) = -\sum_{i=1}^{k} log p(y_t|y_{<t},x,p)
\end{equation}
%\textbf{Coverage Loss} Since the use knowledge is semantically related the conversation, every piece personal knowledge is not used only once. We need to coordinate the semantic relevance and repetition. The loss function only penalize the overlap between each attention distribution and the coverage so far – preventing repeated attention, which is different from the coverage loss function in MT (Machine Translation). The coverage loss is adapted from ... to penalize repeatedly attending to the same personal knowledge:
%\begin{equation}
%    \mathcal{L}(\theta) = \sum_{t=1}^{l_c} \sum_i min(e^t_i,c^t_i)
%\end{equation}

%In summary, the total loss is:
%\begin{equation}
%    \mathcal{L}(\theta)= \mathcal{L}_{NLL}(\theta)+\mathcal{L}_{Con}(\theta)
%\end{equation}
\begin{table*}[htb]
\centering
\resizebox{0.9\textwidth}{!}
{\begin{tabular}{|l|l|c|c|c|}
\hline
\multirow{2}{*}{\textbf{Dataset}}                                    & \multicolumn{1}{l|}{\multirow{2}{*}{\textbf{Model}}} & \multicolumn{3}{c}{\textbf{Automatic Evaluation}}  \\ \cline{3-5} 
                                                            & \multicolumn{1}{c|}{}                  & \textbf{BLEU-1/2} & \textbf{Distinct-1/2} & \textbf{Knowledge R/P/F1} \\ \hline
\multirow{4}{*}{\textbf{ConvAI2(original)}}                      & \textbf{Our model}                              & 0.4118/0.3268  &0.0165/0.0708  & 0.0066/0.0349/0.0112                \\ \cline{2-5} 
                                                            & w/o speaking style                       & 0.3944/0.3117  &0.0079/0.0296  & 0.0058/0.0325/0.0057                \\ \cline{2-5} 
                                                            & w/o knowledge                            & 0.4091/0.3243  &0.0068/0.0253  & 0.0054/0.0287/0.0089                \\ \hline
\multicolumn{1}{|l|}{\multirow{4}{*}{\textbf{ConvAI2(revised)}}} & \textbf{Our model}                              & 0.4163/0.3300  & 0.0081/0.0320  & 0.0041/0.0196/0.0066                \\ \cline{2-5} 
\multicolumn{1}{|l|}{}                                      & w/o speaking style                      & 0.4102/0.3249  &0.0074/0.0282  & 0.0035/0.0176/0.0057                \\ \cline{2-5} 
\multicolumn{1}{|l|}{}                                      & w/o knowledge                            & 0.4065/0.3216  &0.0064/0.0232  & 0.0018/0.0109/0.0057                \\ \hline
\end{tabular}
}
\caption{Ablation experiments with the different models on the persona-chat data.}
\label{ablation evaluation}
\end{table*}
\section{Experiments}
\subsection{Dataset}
We conduct our experiments on two publicly available datasets: CMUDoG \cite{zhou2018dataset} and ConvAI2, which is an extended version of PersonaChat dataset \cite{DBLP:journals/corr/abs-1801-07243}.
The ConvAI2 dataset has separated training and validation set with original and revised persona profiles. The training set contains $131,438$ utterances and there are $7,801$ utterances in valid dataset.

Besides ConvAI2 data, we also experiment with the CMUDoG dataset published in \citet{zhou2018dataset}. The dataset consists of $4,112$ conversations with an average of 21.43 turns and has been divided into the train, valid and test dataset. Distinct from the PersonChat, the CMUDoG is more complex and informative semantic information. Besides, the knowledge in this dataset is about the movie. So the knowledge is more relevant to each other, which is helpful for training our model.

Each dialogue Comprehensive comparisons have been made to the following methods:
\begin{itemize}
    \item Sequence to sequence model with attention \cite{DBLP:journals/corr/VinyalsL15} concats the persona profiles with the history information as inputs. 
    \item The KG-Net, which is proposed by  \citeauthor{DBLP:journals/corr/abs-1902-04911}\shortcite{DBLP:journals/corr/abs-1902-04911} , makes use of both prior and posterior distributions over knowledge to facilitate knowledge selection. The model achieves the state-of-the-art results on the PersonaChat.
\end{itemize}
\subsection{Experiments Settings}
As suggested in \citeauthor{DBLP:journals/corr/abs-1902-04911}\shortcite{DBLP:journals/corr/abs-1902-04911}, we train our model using the following settings. For word embedding representation, we use Glove \cite{pennington2014glove} with an embedding size of 300. For the encoder layer, we use one layer of bidirectional GRU and two different unidirectional GRU for the decoder. And the hidden size of GRU is 800 for ConvAI2 and 500 for CMUDoG. For optimization, we use Adam \cite{kingma2014adam} optimizer with an initial rate of $0.00005$. And to avoid overfitting, we set the dropout rate as 0.3. We clip the gradient when its norm exceeds $5$. And we train our model $25$ epochs or $35$ epochs on the ConvAI2 and CMUDoG respectively.

\begin{table}[htb]
\centering
\resizebox{0.4\textwidth}{!}{
\begin{tabular}{|l|c|c|c|c|}
\cline{1-5}
\multirow{2}{*}{\textbf{Model}} & \multicolumn{4}{c|}{\textbf{repetition}}             \\ \cline{2-5}
                       & \textbf{3}    & \textbf{2}    &\textbf{1}    & \textbf{0}                            \\ \cline{1-5}
Seq2Seq                & 30\% & 8\%  & 8\%  & 54\%                   \\ \cline{1-5}
KG-Net                 & 44\% & 16\% & 16\% & 24\%                   \\ \cline{1-5}
\textbf{Our model}              & 54\% & 9\%  & 20\% & 17\%                   \\ \cline{1-5}
Human                  & 44\% & 35\% & 6\%  & 15\%                    \\ \cline{1-5}
\multirow{2}{*}{}      & \multicolumn{4}{c|}{\textbf{consistency}}            \\ \cline{2-5}
                       & \textbf{3}    & \textbf{2}    &\textbf{1}    & {-} \\ \cline{1-5}
Seq2Seq                & 13\% & 70\% & 17\% & {-} \\ \cline{1-5}
KG-Net                 & 22\% & 60\% & 18\% & {-} \\ \cline{1-5}
\textbf{Our model}              & 23\% & 65\% & 12\% & {-} \\ \cline{1-5}
Human                  & 45\% & 52\% & 3\%  & {-} \\ \cline{1-5}
\end{tabular}
}
\caption{Human evaluation for benchmarks, along with a comparison to human performance.}
	\label{human_evaluation_table}
\end{table}
\section{Result}
\subsection{Automatic Evaluation}
We use BLEU-1/2 \cite{papineni2002bleu}, Distinct-1/2 \cite{li2016diversity} and Knowledge R/P/F1 \cite{dinan2018wizard} to evaluate response generation quality. BLEU-1/2 calculates the average n-gram precision between the generated response and the ground truth. However, because of the one-to-many problems in the dialogue generation, BLEU has poor ability to evaluate the dialogue quality. So we also use Distinct-1/2 to measure the diversity of generated response, which calculates the ratios for unigram and bigram. Besides, we also adopt Knowledge R/P/F1, which is proposed in \citeauthor{dinan2018wizard}\shortcite{dinan2018wizard}, to evaluate how well personal knowledge is expressed. 

Table \ref{automated evaluation} reports evaluation on the ConvAI2 and CMUDoG datasets. As shown in Table \ref{automated evaluation}, our model outperforms most baselines, specifically in Distinct1/2, which means the diversity of generated responses is greatly improved compared to other knowledge-grounded baselines. This verifies our model can not only utilize the personal information to enhance conversation understanding but also provide effective guidance on improving personality understanding based on conversations, which promotes the response with better diversity in turn. Besides, the evaluation of our model on BLEU-1/2 significantly higher than the Seq2Seq model, which demonstrates that our model has the ability to generate high-quality responses.

Besides, compared to the results on the original persona profile setting, the results on the revised persona profile setting don't decrease obviously, while the other baselines have worse performance in most automatic metrics. This means that the other baselines have poor ability to understand complex semantic information of persona profiles and incorporate it into dialogue.

Moreover, the results on the CMUDoG, which has more complex knowledge information and longer history, obviously outperform the other baselines, including the knowledge R/P/F1. There are two reasons. First, our coverage mechanism can help the conversation to select proper and informative knowledge. Second, the interaction module helps the model to incorporate personality-related conversation into the knowledge, resulting in the diversity responses in the subsequent conversation.

\subsection{Ablation Study}
In our model, we take persona profile consistency and speaking style consistency into consideration and design speaking style module and interaction module to enhance dialogue consistency. Different components play different roles. In order to display the necessity and performance of each component, we conduct the ablation experiments on the ConvAI2 dataset with original and revised profiles. From the result shown in Table \ref{ablation evaluation}, we can see that modeling speaking style is greatly helpful for improving dialogue quality and diversity. And it's necessary to consider bilateral information flow of conversation and persona profiles, which contributes more to the improvements in the performance.

\subsection{Human Evaluation}
Since automated metrics are poor for evaluating the repetition and consistency of our model, we also adopt two kinds of human evaluation metrics, which are suggested by \citeauthor{DBLP:journals/corr/abs-1801-07243}\shortcite{DBLP:journals/corr/abs-1801-07243} to evaluate the quality of generated response. Specifically, we selected 100 examples randomly for each model on ConvAI2 with original persona profiles, resulting in 400 examples in total for human evaluation and recruited annotators to rate the consistency, repetition. Note that for each metrics is required to evaluate twice by two participants. The participants are required to score the answers with the following standards.
\begin{itemize}
    \item \textbf{Repetition:} this metric measures whether the generated response tends to use the same personal knowledge multiple times. Score $0$ means response doesn't contain personal knowledge. Score $1$ indicates that response uses the same but unnecessary information. Score $2$ means response use new personal knowledge. Score $3$ represents response contain new and different with personal knowledge. The reason that we set score 3 is that the model will generate some other new topics, which also can be regarded as alleviating repetition problems.
    \item \textbf{Consistency:} we measure the model's ability to generate the consistent response. Specially, $1$ is inconsistent, $2$ is neural and $3$ is consistent. 
\end{itemize}

The human evaluation results are demonstrated in the Table \ref{human_evaluation_table}. We can observe that: (1) our model can avoid repetition by generating more informative responses endowing richer topics than all the baselines, although low ratios in label 2 in repetition demonstrate that our model may be slightly hard to transfer between the given persona profiles. But we observe the dataset and find the fact that the given profiles in the dataset are less relevant to each other. (2) our model can generate more consistent responses than other baselines. In summary, all human judgment results again demonstrate the effectiveness of our model.
\begin{table}[htb]
\resizebox{0.48\textwidth}{!}{
\begin{tabular}{|l|l|}
\hline
Persona   & \begin{tabular}[c]{@{}l@{}}I'm a gold medalist olympian.\\ gymnastics is my favorite sport.\\ my favorite color is yellow.\\ i love italian food and garlic bread.\\ i workout four hours a day.\end{tabular} \\ \hline
Context   & \begin{tabular}[c]{@{}l@{}}A: hello. i love to read and my favorite books \\ \quad are the hunger games series. you?\\ B: hi ! i am an olympic gymnast , i don't \\ \quad read much .\\ A: that is exciting and must allow you to travel\\\quad a lot. ever been to spain ?\\ B: yes , i've . i picked up a lovely yellow\\ \quad \textless{}unk\textgreater there .\\ A: i hope to go back when i retire in a few\\ \quad years . how long before you retire ?\\ B: i've another two years before i can \\ \quad retire from gymnastics .\\ A: what are your plans after retirement ?\end{tabular} \\ \hline
Human     & \begin{tabular}[c]{@{}l@{}} i'd like to go to culinary school ,\\ i am still young. \end{tabular} \\ \hline
Seq2Seq   & i am in the city . i am a student. \\ \hline
KG-Net    & i am a student . i am a bit of a reader. \\ \hline
Our model & i am going to be a doctor . you ?\\ \hline
\end{tabular}
}
\caption{Sampled generation on ConvAI2}
\label{sample}
\end{table}

\subsection{Case Study}
We also evaluate the models by generating the response given the contextual information and personal profiles. Some sampled results are shown in Table \ref{sample}. We observe that our model is capable to generate semantic-related responses comparing the other baselines. Beyond being context-aware, the response also extends the other topics, which make the whole conversation interactive.
\section{Conclusion}
In this paper, we propose a conversation-adaption multi-view persona aware response generation model to keep dialogue consistent and avoid repetition. First, we divide the consistency as the profile consistency and the speaking style consistency. We pay attention to model bilateral information flow of conversation and personal knowledge and design coverage mechanism to avoid repetition. And we introduce speaking style vector and feed it into the decoder to keep speaking style consistency. The experimental evaluation shows that our model can improve the response generation quality. The ablation evaluation verifies the effectiveness of profile consistency and speaking style consistency. The human evaluation shows that our model can avoid repetition and keep consistency compared to the baselines. 

%\section*{Acknowledgments}
%This work is supported by the National Key Research and Development Programs (Grant No.2017YFB0803301).

\section{Acknowledgements}
We would like to thank all of the anonymous reviewers for their invaluable suggestions and helpful comments. This work was supported by the National Natural Science Foundation of China (Grant No 62006222).
\bibliography{aaai21.bib}

\begin{thebibliography}{20}
\providecommand{\natexlab}[1]{#1}
\providecommand{\url}[1]{\texttt{#1}}
\providecommand{\urlprefix}{URL }
\expandafter\ifx\csname urlstyle\endcsname\relax
  \providecommand{\doi}[1]{doi:\discretionary{}{}{}#1}\else
  \providecommand{\doi}{doi:\discretionary{}{}{}\begingroup
  \urlstyle{rm}\Url}\fi

\bibitem[{Bahdanau, Cho, and Bengio(2014)}]{bahdanau2014neural}
Bahdanau, D.; Cho, K.; and Bengio, Y. 2014.
\newblock Neural machine translation by jointly learning to align and
  translate.
\newblock \emph{arXiv preprint arXiv:1409.0473} .

\bibitem[{Bao et~al.(2019)Bao, He, Wang, Lian, and Wu}]{bao2019know}
Bao, S.; He, H.; Wang, F.; Lian, R.; and Wu, H. 2019.
\newblock Know More about Each Other: Evolving Dialogue Strategy via Compound
  Assessment.
\newblock In \emph{Proceedings of the 57th Annual Meeting of the Association
  for Computational Linguistics}, 5382--5391.

\bibitem[{Cho et~al.(2014)Cho, van Merri{\"e}nboer, Gulcehre, Bahdanau,
  Bougares, Schwenk, and Bengio}]{cho-etal-2014-learning}
Cho, K.; van Merri{\"e}nboer, B.; Gulcehre, C.; Bahdanau, D.; Bougares, F.;
  Schwenk, H.; and Bengio, Y. 2014.
\newblock Learning Phrase Representations using {RNN} Encoder{--}Decoder for
  Statistical Machine Translation.
\newblock In \emph{Proceedings of the 2014 Conference on Empirical Methods in
  Natural Language Processing ({EMNLP})}, 1724--1734. Doha, Qatar: Association
  for Computational Linguistics.
\newblock \doi{10.3115/v1/D14-1179}.
\newblock \urlprefix\url{https://www.aclweb.org/anthology/D14-1179}.

\bibitem[{Dinan et~al.(2018)Dinan, Roller, Shuster, Fan, Auli, and
  Weston}]{dinan2018wizard}
Dinan, E.; Roller, S.; Shuster, K.; Fan, A.; Auli, M.; and Weston, J. 2018.
\newblock Wizard of wikipedia: Knowledge-powered conversational agents.
\newblock \emph{arXiv preprint arXiv:1811.01241} .

\bibitem[{Kingma and Ba(2014)}]{kingma2014adam}
Kingma, D.~P.; and Ba, J. 2014.
\newblock Adam: A method for stochastic optimization.
\newblock \emph{arXiv preprint arXiv:1412.6980} .

\bibitem[{Li et~al.(2016{\natexlab{a}})Li, Galley, Brockett, Gao, and
  Dolan}]{li2016diversity}
Li, J.; Galley, M.; Brockett, C.; Gao, J.; and Dolan, W.~B. 2016{\natexlab{a}}.
\newblock A Diversity-Promoting Objective Function for Neural Conversation
  Models.
\newblock In \emph{Proceedings of the 2016 Conference of the North American
  Chapter of the Association for Computational Linguistics: Human Language
  Technologies}, 110--119.

\bibitem[{Li et~al.(2016{\natexlab{b}})Li, Monroe, Ritter, Jurafsky, Galley,
  and Gao}]{li2016deep}
Li, J.; Monroe, W.; Ritter, A.; Jurafsky, D.; Galley, M.; and Gao, J.
  2016{\natexlab{b}}.
\newblock Deep Reinforcement Learning for Dialogue Generation.
\newblock In \emph{Proceedings of the 2016 Conference on Empirical Methods in
  Natural Language Processing}, 1192--1202.

\bibitem[{Lian et~al.(2019)Lian, Xie, Wang, Peng, and
  Wu}]{DBLP:journals/corr/abs-1902-04911}
Lian, R.; Xie, M.; Wang, F.; Peng, J.; and Wu, H. 2019.
\newblock Learning to Select Knowledge for Response Generation in Dialog
  Systems.
\newblock \emph{CoRR} abs/1902.04911.
\newblock \urlprefix\url{http://arxiv.org/abs/1902.04911}.

\bibitem[{Luan et~al.(2017)Luan, Brockett, Dolan, Gao, and
  Galley}]{DBLP:journals/corr/abs-1710-07388}
Luan, Y.; Brockett, C.; Dolan, B.; Gao, J.; and Galley, M. 2017.
\newblock Multi-Task Learning for Speaker-Role Adaptation in Neural
  Conversation Models.
\newblock \emph{CoRR} abs/1710.07388.
\newblock \urlprefix\url{http://arxiv.org/abs/1710.07388}.

\bibitem[{Papineni et~al.(2002)Papineni, Roukos, Ward, and
  Zhu}]{papineni2002bleu}
Papineni, K.; Roukos, S.; Ward, T.; and Zhu, W.-J. 2002.
\newblock Bleu: a method for automatic evaluation of machine translation.
\newblock In \emph{Proceedings of the 40th annual meeting of the Association
  for Computational Linguistics}, 311--318.

\bibitem[{Pennington, Socher, and Manning(2014)}]{pennington2014glove}
Pennington, J.; Socher, R.; and Manning, C.~D. 2014.
\newblock GloVe: Global Vectors for Word Representation.
\newblock In \emph{Empirical Methods in Natural Language Processing (EMNLP)},
  1532--1543.
\newblock \urlprefix\url{http://www.aclweb.org/anthology/D14-1162}.

\bibitem[{Qian et~al.(2017)Qian, Huang, Zhao, Xu, and
  Zhu}]{DBLP:journals/corr/QianHZXZ17}
Qian, Q.; Huang, M.; Zhao, H.; Xu, J.; and Zhu, X. 2017.
\newblock Assigning personality/identity to a chatting machine for coherent
  conversation generation.
\newblock \emph{CoRR} abs/1706.02861.
\newblock \urlprefix\url{http://arxiv.org/abs/1706.02861}.

\bibitem[{See et~al.(2019)See, Roller, Kiela, and Weston}]{see2019makes}
See, A.; Roller, S.; Kiela, D.; and Weston, J. 2019.
\newblock What makes a good conversation? how controllable attributes affect
  human judgments.
\newblock \emph{arXiv preprint arXiv:1902.08654} .

\bibitem[{Serban et~al.(2016)Serban, Lowe, Charlin, and
  Pineau}]{DBLP:journals/corr/SerbanLCP16}
Serban, I.~V.; Lowe, R.; Charlin, L.; and Pineau, J. 2016.
\newblock Generative Deep Neural Networks for Dialogue: {A} Short Review.
\newblock \emph{CoRR} abs/1611.06216.
\newblock \urlprefix\url{http://arxiv.org/abs/1611.06216}.

\bibitem[{Shang, Lu, and Li(2015)}]{shang2015neural}
Shang, L.; Lu, Z.; and Li, H. 2015.
\newblock Neural responding machine for short-text conversation.
\newblock \emph{arXiv preprint arXiv:1503.02364} .

\bibitem[{Sutskever, Vinyals, and Le(2014)}]{sutskever2014sequence}
Sutskever, I.; Vinyals, O.; and Le, Q.~V. 2014.
\newblock Sequence to sequence learning with neural networks.
\newblock \emph{arXiv preprint arXiv:1409.3215} .

\bibitem[{Vinyals and Le(2015)}]{DBLP:journals/corr/VinyalsL15}
Vinyals, O.; and Le, Q.~V. 2015.
\newblock A Neural Conversational Model.
\newblock \emph{CoRR} abs/1506.05869.
\newblock \urlprefix\url{http://arxiv.org/abs/1506.05869}.

\bibitem[{Xing et~al.(2018)Xing, Wu, Wu, Huang, and
  Zhou}]{DBLP:conf/aaai/XingWWHZ18}
Xing, C.; Wu, Y.; Wu, W.; Huang, Y.; and Zhou, M. 2018.
\newblock Hierarchical Recurrent Attention Network for Response Generation.
\newblock In \emph{Proceedings of the Thirty-Second {AAAI} Conference on
  Artificial Intelligence, (AAAI-18), the 30th innovative Applications of
  Artificial Intelligence (IAAI-18), and the 8th {AAAI} Symposium on
  Educational Advances in Artificial Intelligence (EAAI-18), New Orleans,
  Louisiana, USA, February 2-7, 2018}, 5610--5617.
\newblock
  \urlprefix\url{https://www.aaai.org/ocs/index.php/AAAI/AAAI18/paper/view/16510}.

\bibitem[{Zhang et~al.(2018)Zhang, Dinan, Urbanek, Szlam, Kiela, and
  Weston}]{DBLP:journals/corr/abs-1801-07243}
Zhang, S.; Dinan, E.; Urbanek, J.; Szlam, A.; Kiela, D.; and Weston, J. 2018.
\newblock Personalizing Dialogue Agents: {I} have a dog, do you have pets too?
\newblock \emph{CoRR} abs/1801.07243.
\newblock \urlprefix\url{http://arxiv.org/abs/1801.07243}.

\bibitem[{Zhou, Prabhumoye, and Black(2018)}]{zhou2018dataset}
Zhou, K.; Prabhumoye, S.; and Black, A.~W. 2018.
\newblock A Dataset for Document Grounded Conversations.
\newblock In \emph{Proceedings of the 2018 Conference on Empirical Methods in
  Natural Language Processing}, 708--713.

\end{thebibliography}

\end{document}